\def\year{2020}\relax
\documentclass[letterpaper]{article} 
\usepackage{aaai20}  
\usepackage{times}  
\usepackage{helvet} 
\usepackage{courier}  
\usepackage[hyphens]{url}  
\usepackage{graphicx} 
\usepackage{graphicx}  
\nocopyright
\usepackage{amsmath,amssymb,amsfonts}
\usepackage{xcolor}
\usepackage{booktabs}
\usepackage{threeparttable}

\title{Extended Answer and Uncertainty Aware Neural Question Generation}

\author{Hongwei Zeng\thanks{These two authors contributed equally.}$^{1,2}$, Zhuo Zhi$^{*3}$, Jun Liu$^{1,2}$, Bifan Wei$^{2,4}$ \\
	$^{1}$School of Electronic and Information Engineering, Xi'an Jiaotong University \\
	$^{2}$National Engineering Lab for Big Data Analytics, Xi'an Jiaotong University \\
	$^{3}$School of Electrical Engineering, Xi'an Jiaotong University \\
	$^{4}$School of Continuing Education, Xi'an Jiaotong University \\
	\texttt{\{zhw1025@stu., zhizz001@stu., liukeen@, bifanwei@\}xjtu.edu.cn}
	\\}
 
\begin{document}

\maketitle

\begin{abstract} 
In this paper, we study automatic question generation, the task of creating questions from corresponding text passages where some certain spans of the text can serve as the answers. 
We propose an Extended Answer-aware Network (EAN) which is trained with Word-based Coverage Mechanism (WCM) and decodes with Uncertainty-aware Beam Search (UBS).
The EAN represents the target answer by its surrounding sentence with an encoder, 
and incorporates the information of the extended answer into paragraph representation with gated paragraph-to-answer attention to tackle the problem of the inadequate representation of the target answer.
To reduce undesirable repetition,
the WCM penalizes repeatedly attending to the same words at different time-steps in the training stage.
The UBS aims to seek a better balance between the model confidence in copying words from an input text paragraph and the confidence in generating words from a vocabulary.
We conduct experiments on the SQuAD dataset, and the results show our approach achieves significantly performance improvement.
\end{abstract}

\section{Introduction}
Question generation (QG) aims to automatically generate questions from corresponding natural language text passages.
As a challenging and complementary task to question answering (QA), QG has received increasing attention in various kind of applications in recent years. 
In the field of education, QG can help with reading practice and assessment since generating question manually is time-consuming \cite{heilman-smith-2010-good}. 
In conversational systems and chatbots, such as Siri, Cortana, and Google Assistant, QG can serve as an important component to start a conversation or request feedback \cite{mostafazadeh-etal-2016-generating}.
QG can also be utilized to generate large-scale question-answer corpus to augment training data for other tasks, such as machine reading comprehension and QA, and to assist in improving the performance of their models \cite{du-etal-2017-learning}. 

Compared with previous works for QG which utilized heuristic rules to transform the declarative sentences into interrogative questions \cite{chali2015towards,heilman2011automatic}, 
recent neural network-based models are fully data-driven and do not rely on manually designed rules. 
The neural network-based models utilize a sequence to sequence (seq2seq) model with attention mechanism to deal with the question generation problem in an end-to-end fashion. 

However, existing neural question generation models still suffer two issues.
One is that the models have insufficient learning of the representation of the target answer. 
\cite{A-Preliminary-Study,zhao2018paragraph} only utilize the answer positions for sentence or paragraph encoding, which cannot learn an independent and explicit representation for the target answer.
\cite{song2018unified,kim2019improving} introduced another answer encoder separately for answer encoding, which will bring the increment of model parameters and training difficulty. 
Furthermore, the average number of the target answer words is only 3.4 on the SQuAD dataset, 
and the answer words belonging to the type of Numeric or Person \cite{rajpurkar-etal-2016-squad} are likely to be replaced by a generic $<$\textit{UNK}$>$ token.
The lack of context will result in incomplete and inaccurate representation. 
The other one is the repetition problem.
\cite{tu-etal-2016-modeling,see-etal-2017-get} utilize the coverage mechanism to penalize repeatedly attending to the same locations. 
However, if a word appears multiple times in different locations, it is still likely to be copied multiple times with little penalization. 
\cite{zhao2018paragraph} proposed a maxout pointer to limit the copy scores of repeated words to their maximum value at each time-step. 
However, this method completely ignores the fact that the words repeatedly appearing in the input passage may suggest its higher importance and even necessary repetition in the generated question. 

Besides, we have observed the inconsistent phenomenon between the copy probability and the corresponding copy distribution in the copy mechanism. 
More specifically, 
there are many words copied from the paragraph with high copy probability but very flat copy distribution. 
The flat distribution refers to high uncertainty according to the definition of the information entropy.
That is to say, the model will tend to copy a word from the input paragraph with high copy probability, despite the model has low confidence of which word to copy.

In this paper, 
we propose an Extended Answer-aware Network (EAN) which is trained with Word-based Coverage Mechanism (WCM) and decodes with Uncertainty-aware Beam Search (UBS).
The EAN contains an extended answer-aware encoder and a decoder with attention, copy and coverage mechanism.
Our encoder treats the paragraph and the extended answer separately to better utilize the information from both sides. 
Instead of utilizing the existing answer which contains average only 3.4 words, 
we extend the answer representation by its surrounding sentence (much shorter than the paragraph) 
and learn a more comprehensive representation of the target answer together with its context. 
Besides, paragraph encoder and answer encoder share the same parameters. 
Then, the gated paragraph-to-answer attention is designed to incorporate the extended answer information into paragraph representation. 
The attention-based decoder with copy mechanism is finally utilized to generate the target question. 
In the training stage, we design a word-based coverage mechanism to penalize repeatedly attending to the same words, 
which will reduce repetition caused by repeatedly attending to the same words at different time-steps.
In the decoding stage, we incorporate a well-designed uncertainty score into beam search via linear combination,
which seek a better balance between the confidence of the model to copy words from the input paragraph or to generate words from the vocabulary.
We conduct extensive experiments on SQuAD, and the experiment results show that our approach achieves significant improvement over other baselines.

\section{Related Work}
\noindent \textbf{Question Generation} 
Existing QG approaches can be mainly classified into two categories: rule-based approaches and neural network-based approaches. 
The traditional rule-based QG approaches utilized well-designed rules and templates to transform the declarative sentences into interrogative questions \cite{chali2015towards,heilman2011automatic}. 
However, these approaches heavily rely on hand-crafted rules and templates designed by linguistic experts which is extremely expensive. 

Recently, the neural network-based approaches train end-to-end neural networks from scratch.  
\cite{du-etal-2017-learning} first tackle the QG using the attention-based encoder-decoder framework in an end-to-end fashion. 
However, their model does not consider the target answer, resulting in randomly generated questions.
There are mainly two ways to incorporate the target answer information.
One is to encode the answer word location using an annotation vector as additional word features.
\cite{A-Preliminary-Study} took rich features including the answer positions as the input at the sentence level. 
\cite{zhao2018paragraph} proposed a gated self-attention encoder to effectively utilize relevant information with answer tagging at the paragraph level. 
The other one is to employ another answer encoder for the target answer encoding.
\cite{song2018unified} utilized the multi-perspective matching encoder to perform comprehensive understanding between the target answer and the passage.
\cite{kim2019improving} proposed an answer-separated seq2seq model which treats the target answer and the passage separately to better utilize the information from both sides.

Additionally, copy mechanism \cite{gulcehre-etal-2016-pointing} or pointer network \cite{see-etal-2017-get} was introduced to allow copying words from input paragraph via pointing.
\cite{zhao2018paragraph} designed a maxout pointer mechanism to limits the magnitude of copy scores of repeated words to their maximum value, which reduces the repetition issue brought by the basic copy mechanism. 
\cite{liu2019learning} trained a clue word predictor to identify whether each word in the input paragraph is a clue word that may be copied into the target question and guide the model to learn accurate boundaries between copying words and generating words from the decoder vocabulary.

\noindent \textbf{Beam Search} 
Beam search is a heuristic search algorithm that explores a graph by expanding B most promising nodes at each time-step, where B is called the beam-width.
The successful utilization of beam search has led to significant improvements for many language generation tasks such as Neural Machine Translation (NMT) and image captioning. 
\cite{DBLP:journals/corr/LiJ16} proposed a diversification heuristic for beam search to discourages sequences from sharing common roots, implicitly resulting in diverse lists.
\cite{li-etal-2018-simple} incorporates a coverage metric into beam search to address the problem of over-translation and under-translation in NMT. 
\cite{Diverse-Beam-Search} incorporates diverse constraints into beam search which results in improvements on both oracle task-specific and diversity-related metrics for image captioning.

\section{Problem Statement}
In this section, we define the problem of question generation. Given a paragraph sequence $ X^p = (x^p_1, \dots, x^p_{n_p}) $, 
a target answer sequence $ X^a = (x^a_1, \dots, x^a_{n_a}) $, 
and a sentence sequence $ X^s = (x^s_1, \dots, x^s_{n_s}) $ which is a sub span of $X^p$ and contains $X^a$.
We term the sentence $X^s$ as the extended answer of $X^a$.
Our goal is to generate the question sequence $ Y = (y_1, \dots, y_T) $ based on the information of $ X^p, X^s, X^a $. That is, the task is to generate $ \hat{Y}$ such that:
\begin{equation}\label{conp}
\begin{split}
\hat{Y} & = \mathop{\arg\max}_{Y} P_\theta(Y|X^p, X^s, X^a) \\
		& = \mathop{\arg\max}_{Y} \sum_{t=1}^{T} P_\theta(y_t|X^p, X^s, X^a, y_{<t})
\end{split}
\end{equation}
where $P_\theta(y_t|X^p, X^s, X^a, y_{<t})$ is abbreviated as $P_\theta(y_t|y_{<t})$ later for simplicity.
The parameter $\theta$ is optimized by maximum likelihood estimation.

\section{Model}\label{model}

\begin{figure*}[ht]
	\centerline{\includegraphics[width = 1\textwidth]{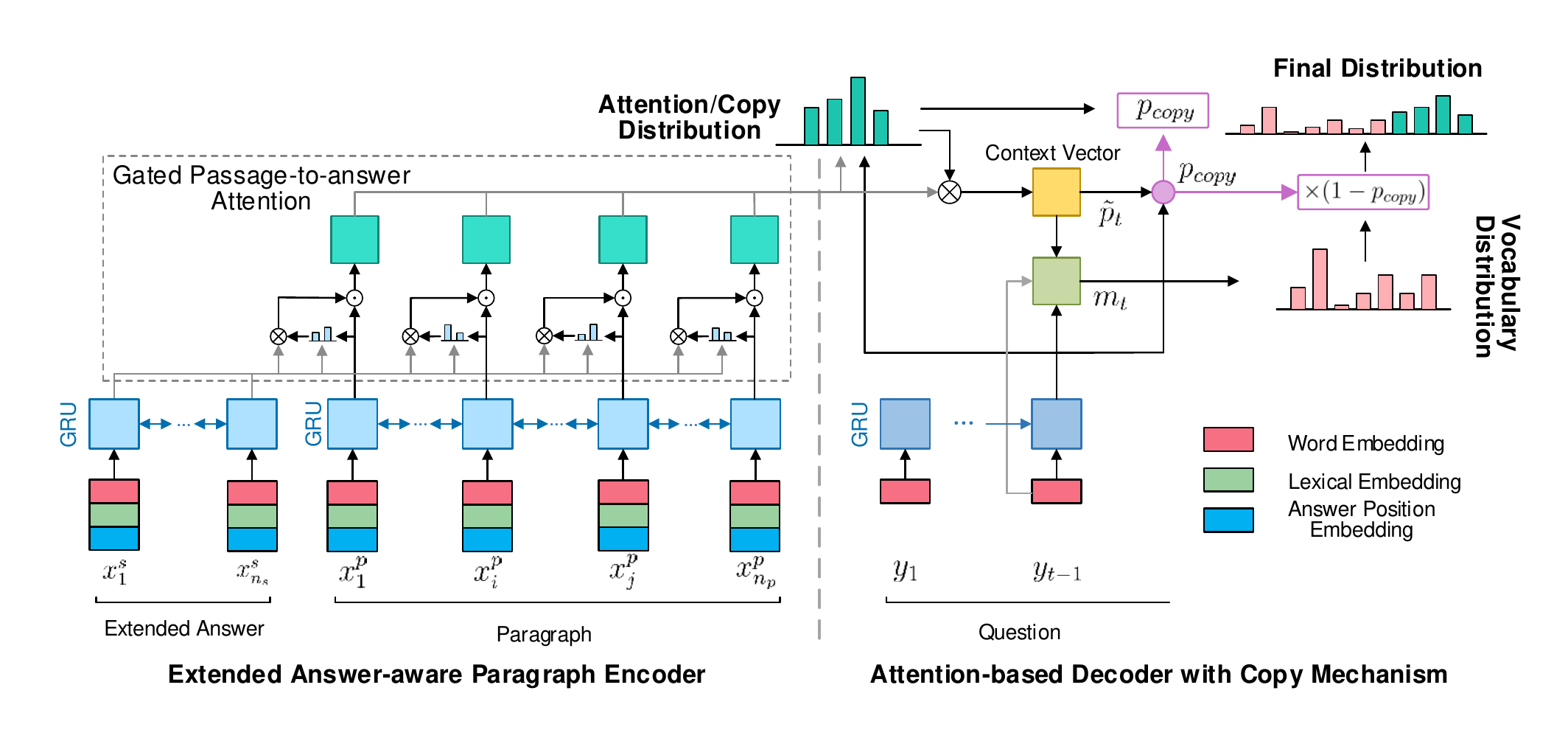}}
	\caption{Architecture diagram of our proposed EAN containing an Extended Answer-aware Paragraph Encoder and an Attention-based Decoder with Copy Mechanism.}
	\label{model_diagram}
\end{figure*}

In this section, we introduce our proposed model in detail.
The novel components of our model mainly consist of an extended answer-aware encoder (EAN) taking as input the whole paragraph and its related features while attending the extended answer, 
and a word-based coverage mechanism (WCM) penalizing repeatedly attending to the same words at different time-steps. 
During decoding, our model decodes with uncertainty-aware beam search (UBS), which incorporates a well-designed uncertainty score into beam search to seek higher quality questions. 
Figure~\ref{model_diagram} illustrates the overall architecture of our proposed model.

\subsection{Extended Answer-aware Paragraph Encoder}\label{encoder}
\noindent \textbf{Paragraph Encoder} We first represent the paragraph-level joint embeddings of words, answer tags and lexical features including NER and POS as $W^p = (w^p_1, \dots, w^p_{n_p})$, 
where $n_p$ is the length of the paragraph $X^p$, $w^p_i \in \mathbb{R}^{d_w+d_a+d_n+d_p} $, and $d_w, d_a, d_n, d_p$ is the dimensionality of word embedding, answer position embedding, NER embedding and POS embedding respectively. 
To capture more context information, we utilize a bi-directional GRU (BiGRU) to encode the paragraph which takes $W^p$ as input and produces the forward and backward hidden states $\overrightarrow{P} = (\overrightarrow{p}_{1}, \dots, \overrightarrow{p}_{n_p})$, $\overleftarrow{P} = (\overleftarrow{p}_{1}, \dots, \overleftarrow{p}_{n_p})$. 
Then they are concatenated to form the paragraph representation $P = (p_1, \dots, p_{n_p})$. This feature-rich encoding process for time-step $i$ of the paragraph is calculated as follows:
\begin{equation}
\overrightarrow{p}_i = \textbf{BiGRU}(w^p_i, \overrightarrow{p}_{i-1})
\end{equation}
\begin{equation}
\overleftarrow{p}_i = \textbf{BiGRU}(w^p_i, \overleftarrow{p}_{i-1})
\end{equation}
\begin{equation}
p_i = [\overrightarrow{p}_i; \overleftarrow{p}_i ]
\end{equation}

\noindent \textbf{Extended Answer Encoder} 
Rather than feeding the answer encoder with the short answer span $X^a$, 
we use the sentence $X^s$ in which the answer is located to replace the original short answer span to provide more informative answer information.
Likewise, we represent the sentence surrounding the target answer $X^s$ as $W^s = (w^s_1, \dots, w^s_{n_s})$ using the sentence-level joint embeddings, 
where $n_s$ is the length of the sentence $X^s$.
Then we use the same BiGRU to encode the sentence to get $S = (s_1, \dots, s_{n_s})$. 
Such operation helps the model capture more context-aware answer information without increasing the number of parameters. 
The $i$ time-step of the extended answer representation is calculated as follows:
\begin{equation}
\overrightarrow{s}_i = \textbf{BiGRU}(w_i, \overrightarrow{s}_{i-1})
\end{equation}
\begin{equation}
\overleftarrow{s}_i = \textbf{BiGRU}(w_i, \overleftarrow{s}_{i-1})
\end{equation}
\begin{equation}
s_i = [\overrightarrow{s}_i; \overleftarrow{s}_i ]
\end{equation}

\noindent \textbf{Gated Paragraph-to-answer Attention} 
We propose a gated paragraph-to-answer attention to incorporate the extended answer information into paragraph representation and to determine the importance of the information regarding the extended answer in the paragraph.
We first calculate the attention-pooling vector of the whole extended answer $S$ for $p_j$ as follows:
\begin{equation}\label{att}
\alpha^e_{j} = softmax(S^\mathsf{T}\mathbf{W}_sp_j)
\end{equation}
\begin{equation}\label{sr}
\tilde{s}_j = \sum_{i=1}^{n_s}\alpha^e_{ji} s_i
\end{equation}
where $\alpha^e_j \in \mathbb{R}^{n_s}$ is the attention distribution, and $\mathbf{W}_s$ is trainable parameters to be learned.
Then we combine the original paragraph representation $p_j$ with the attended answer representation $\tilde{s}_j$ to produce the fused representation $f_j$. 
An additional gate is utilized to select the information between $p_j$  and $f_j$.
\begin{equation}\label{cb}
f_j = \tanh (\mathbf{W}_f [p_j; \tilde{s}_j])
\end{equation}
\begin{equation}\label{gt}
g_j = sigmoid(\mathbf{W}_g [p_j; \tilde{s}_j])
\end{equation}
\begin{equation}\label{rp}
\hat{p}_j = g_j \circ p_j + (1 - g_j) \circ f_j
\end{equation}
where $g_j$ is a learnable gate vector and $\circ$ is a element-wise multiplication operator. 
The gate effectively models the phenomenon that only parts of the paragraph are relevant for the target question.

\subsection{Attention-based Decoder with Copy Mechanism}\label{decoder}
We employ another GRU as the decoder to generate question words sequentially conditioned on the encoded input information and the previously decoded words. 
The hidden state of the decoder is initialized as follow:
\begin{equation}
h_0 = tanh(\mathbf{W}_0 \overleftarrow{p}_1 + b)
\end{equation}
where $\overleftarrow{p}_{1}$ is the last backward paragraph encoder hidden state.

At each decoding step $t$, the GRU decoder takes as input the previous word embedding $w^y_{t-1}$, and context vector $\tilde{p}_{t-1}$ to compute the new hidden state $h_t$.
\begin{equation}\label{eq:d1}
h_t = \textbf{GRU}(h_{t-1}, [w^y_{t-1}; \tilde{p}_{t-1}])
\end{equation}

The context vector $\tilde{p}_{t}$ for current time-step $t$ is computed through the concatenate attention mechanism  \cite{luong-etal-2015-effective} as follow:
\begin{equation}\label{eq:d2}
e_{t, i} = v^\mathsf{T} \tanh(\mathbf{W}_h h_t + \mathbf{W}_p \hat{p_i})
\end{equation}
\begin{equation}\label{eq:d3}
\alpha^d_{t, i} = \frac{\exp(e_{t,i})}{ \sum_{j=1}^{n_p} \exp(e_{t,j})}
\end{equation}
\begin{equation}\label{eq:d4}
\tilde{p}_{t} = \sum_{i=1}^{n_p} \alpha^d_{t,i} \hat{p_i}
\end{equation}
where $e_{t, i}$ is the importance score matching the current decoder state $h_t$ with each encoded paragraph representation $\hat{p_i}$, and $\alpha^d_{t, i}$ is the normalized attention weight on the encoded paragraph representation $\hat{p_i}$ at current time-step $t$.

Utilizing the previous word embedding $w^y_{t-1}$, the current context vector $\tilde{p}_{t}$, and the current decoder state $h_t$ we compute the readout state $r_t$ which is then passed through a maxout hidden layer  \cite{Goodfellow2013MaxoutN} to obtain the probability distribution of the next word over the decoder vocabulary with a softmax layer.
\begin{equation}\label{eq:d5}
r_t = \mathbf{W}_r w^y_{t-1} + \mathbf{U}_r \tilde{p}_{t} + \mathbf{V}_r h_t
\end{equation}
\begin{equation}\label{eq:d6}
m_t = [\mathop{\max}\{r_{t, 2j-1}, r_{t, 2j}\}]^\mathsf{T}_{j=1, \dots, d}
\end{equation}
\begin{equation}\label{eq:d7}
P_{vocab}(y_t|y_{<t}) = softmax(\mathbf{W}_m m_t)
\end{equation}

\noindent \textbf{Copy Mechanism}
Different from the copy mechanism in \cite{A-Preliminary-Study}, 
we label a target question word as a word copied from the paragraph input as long as it appears in both the paragraph input and the target question. 
This helps us deal with not only the rare and unknown words problem, but also reproduce factual details more accurately. 
The copy mechanism takes as input the current decoder state $h_t$ and the context vector $\tilde{p}_{t}$ to produce the probability $P_{c}$ of copying a word from the paragraph:
\begin{equation}\label{eq:copy_probability}
P_{c} = \sigma (\mathbf{W}_{ch} h_t + \mathbf{W}_{cp} \tilde{p}_t + b)
\end{equation}
where $\sigma$ is the sigmoid function. 
We construct another vocabulary $\mathcal{X}$ for all the \textit{unique} words in paragraph $X^p$.
Then the copy probability over the words of paragraph can be calculated as follow:
\begin{equation} \label{na}
P_{copy}(y_t|y_{<t}) = \sum_{i, where\ x^p_i=y_t}\alpha^d_{t, i}
\end{equation}

Therefore, the probability distribution of the next word $y_t$ in Eq. (\ref{eq:d7}) can be replaced with:
\begin{equation}\label{eq:final_probability}
P(y_t|y_{<t}) = (1 - P_{c}) P_{vocab}(y_t|y_{<t}) + P_{c} P_{copy}(y_t|y_{<t})
\end{equation}

\subsection{Training with Word-based Coverage Mechanism} 
It is observed that the repetition issue becomes more severe when the encoder takes paragraph as input  \cite{zhao2018paragraph}. To alleviate this problem, we maintain a coverage vector $c_t$ to record the degree of coverage that each word has received from the attention mechanism thus far.
\begin{equation}
c_t = \sum_{t^\prime = 1}^{t-1} a^d_{t^\prime}
\end{equation}
where $c_t$ is simply the sum of attention distributions over all previous decoder time-steps, and $c_0$ is initialized as a zero vector. $c_t$ is then used to inform the attention mechanism by replacing Eq. (\ref{eq:d2}) with:
\begin{equation}\label{eq:d1}
e_{t, i} = v^\mathsf{T} \tanh(\mathbf{W}_h h_t + \mathbf{W}_p \hat{p_i} + \mathbf{W}_c c_{t, i})
\end{equation}

Instead of using the attention weights and the coverage scores over the positions of paragraph, we introduce a modified coverage loss to penalize repeatedly attending to the same words:
\begin{equation}
\mathcal{L}^{cov}_t =  \sum_{k \in \mathcal{X}} \mathop{\min} (P_{copy}(k|y_{<t}), \tilde{c}_{t}(k))
\end{equation}
\begin{equation}
\tilde{c}_{t}(k) = \sum_{i, where\ x^p_i=k} c_{t, i}
\end{equation}
where $\tilde{c}_{t}(k)$ are the sum of ${c}_{t,i} $ on the positions where the encoder GRU takes the word $k$ as input.
The coverage loss and the primary loss are combined to form the final loss function:
\begin{equation} \label{cov}
\mathcal{L}_t = -\mathrm{log} P(y_t|y_{<t}) + \lambda \sum_{k \in \mathcal{X}} \mathop{\min} (P_{copy}(k|y_{<t}), \tilde{c}_{t}(k))
\end{equation}
where $\lambda$ is the hyperparameter to reweight the coverage loss.

\subsection{Decoding with Uncertainty-aware Beam Search}\label{iebs}
\begin{figure}[h]
	\centerline{\includegraphics[width = 0.95\columnwidth]{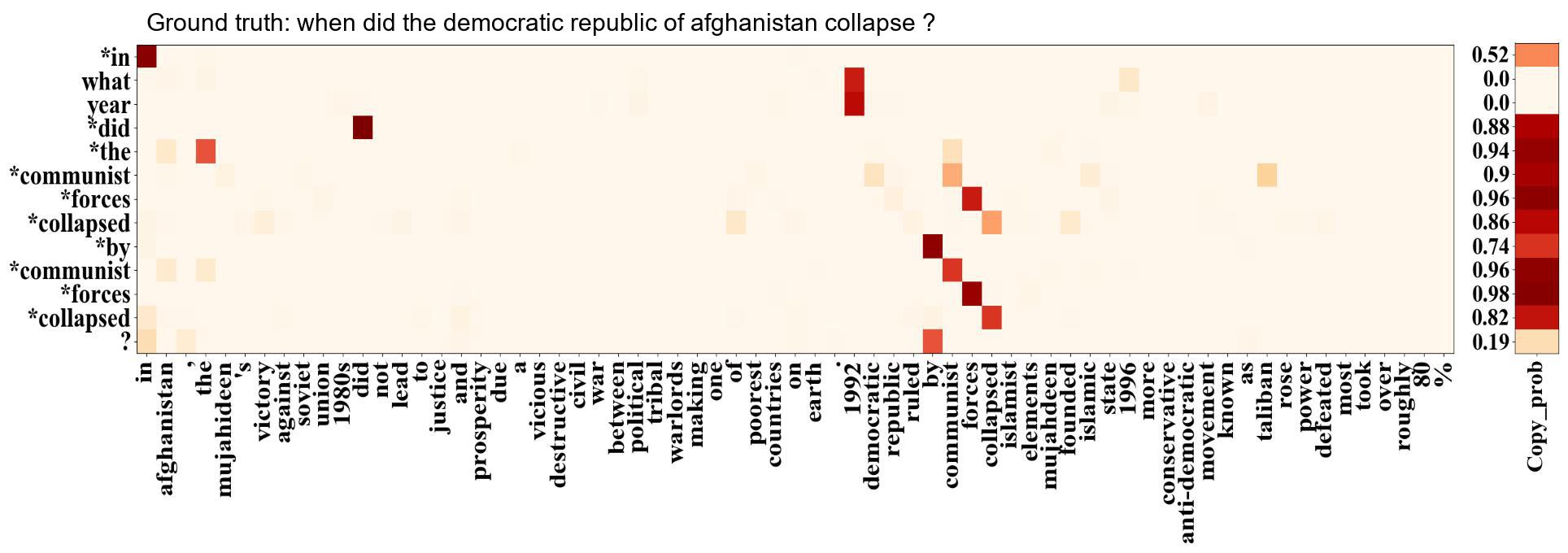}}
	\caption{Copy heatmap: each row represents a copy distribution and a corresponding copy probability decoded by model EAN+WCM.
		The words marked with * are words copied from input paragraph.}
	\label{nie}
\end{figure}

According to the observation, there is inconsistency between the copy probability $P_c$ deciding whether to copy or not and the copy distribution $P_{copy}$ deciding which word to be copied in the copy mechanism.
Theoretically, 
the higher the copy probability $P_c$ is, the more likely the model will be to copy a word from the paragraph. 
Besides, the more flat the copy distribution $P_{copy}$ is, the higher the uncertainty of knowing which word to be copied will be.
However, these two are not guaranteed to show a consistent tendency.
For example, there are repetitions of words, ``communist'', ``forces'' and ``collapsed'', which have been generated twice all with high copy probabilities (as depicted in Figure~\ref{nie}).
These three words are copied with flat distribution at the first time, which means that the model is uncertain about which word to copy.
This may result in a copy of the undesired word.
The inconsistent phenomenon between the probability of generating words from vocabulary $1 - P_c$ and the vocabulary distribution $P_{vocab}$ also exists.

\noindent \textbf{Uncertainty Score} 
To address the problem discussed above, 
we introduce an uncertainty score $u_t$ which takes a comprehensive consideration of the copy probability and the information entropy of the vocabulary distribution and the copy distribution at time-step $t$: 
\begin{equation}
u_t = (1 - P_c)\frac{H[P_{vocab}( y_t |y_{<t})]}{\mathrm{log}\ |\mathcal{V}|} + P_c\frac{H[P_{copy}( y_t |y_{<t})]}{\mathrm{log}\ |\mathcal{X}|}
\end{equation}
where $H [P(X)] = \mathbb{E} [-\mathrm{log}(P(X))] $ is the information entropy of the distribution $P(X)$. 
The information entropy can refer to the measurement of uncertainty, and $|\mathcal{V}|$, $|\mathcal{X}|$ are the size of the vocabulary and the number of the unique words in the paragraph respectively. 
Thus, at decoding time-step $t$, the score $u_t$ reveals the uncertainty of the model to generate the next word $y_t$ from both the decoder vocabulary $\mathcal{V}$ and the copy vocabulary $\mathcal{X}$.
The larger $u_t$ is, the higher uncertainty of the model to generate the word $y_t$ will be.

We then incorporate the uncertainty score $u_t$ into beam search via linear combination with the conditional log-likelihood of the model as follow:
\begin{equation} \label{beam}
s(y_{1:T'}) = (1 - \beta) \frac{1}{T'}\sum_{t=1}^{T'} \mathrm{log} P(y_{t}|y_{<{t}}) + \beta \log(\frac{1}{\frac{1}{T'}\sum_{t=1}^{T'} u_t})
\end{equation}
where $y_{1:T'}$ is the partially generated question, $\beta$ is the hyperparameter for linear interpolation, and $s(y_{1:T'})$ is the scoring function which are employed for the ranking of candidate questions in beam search.

\section{Experiments}
In this section, we evaluate the performance of the proposed model in comparison with several baselines in the QG dataset. The experimental results indicate that our method can effectively capture more relevant contextual information, and reduce undesired repetition, and improve the performance of the existing beam search.

\begin{table*}[ht]
	\centering
	\caption{Performace of Different Models on SQuAD Dataset}
	\label{tab_scores}
	\begin{threeparttable}
		\begin{tabular}{c|c|c|c|c|c|c}
			\toprule
			Models      & BLEU-1 & BLEU-2 & BLEU-3 & BLEU-4 & METEOR & ROUGE-L \\ \midrule
			NQG++*    & 42.60   & 26.50   & 18.49   & 13.40  & 18.47  & 41.55   \\
			MPQG 								& -			& -		& - 		& 13.91 & - 		& -		\\
			ASs2s 								& -			& -		& - 		& 16.17 & - 		& -		\\
			s2s-a-at-mp-gsa 	& 45.69 	& 30.25 & 22.16 	& 16.85 & 20.62 	& 44.99 \\
			CGC-QG &46.58 &30.90 &22.82 &17.55 &21.24 &44.53   \\ 
			\midrule
			gsa-WCM   & 47.05   & 30.90   & 22.50   & 16.97  & 21.33   & 44.70   \\
			EAN  & 46.88   & 31.01   & 22.71   & 17.20   & 21.20  & 44.75   \\
			EAN-WCM      & 47.21   & 31.35   & 22.99   & 17.41  & 21.43  & 44.98 \\
			EAN-WCM-UBS    & \textbf{47.25}   & \textbf{31.46}   & \textbf{23.13}   & \textbf{17.56}   & \textbf{21.51}  & \textbf{45.00}   \\ \bottomrule
		\end{tabular}
		\begin{tablenotes}
			\footnotesize
			\item The baseline marked with ``*'' is conducted using released source code. The unreported metrics are marked with ``-".
		\end{tablenotes}
	\end{threeparttable}
\end{table*}

\subsection{Dataset}
We conduct experiments on the SQuAD 1.1 \cite{rajpurkar-etal-2016-squad} dataset which is a reading comprehension dataset consisting of questions posed by crowd-workers on a set of Wikipedia articles covering a wide range of topics. The original SQuAD dataset contains 23,215 paragraphs from 536 articles with over 100,000+ question-answer pairs where the answer to each question is a span of tokens in the corresponding reading passage. Since the test set is not publicly available, we follow the data split proposed by  \cite{A-Preliminary-Study} where the original dev set is randomly split into dev and test sets with ratio 50\%-50\%. We extract paragraph-sentence-answer-question quadruplets to build the training, development and test sets. The Standford CoreNLP toolkit  \cite{manning-etal-2014-stanford} is used to annotate POS and NER tags in the dataset.

\subsection{Metrics}
We evaluate the performance of our model with the following evaluation metrics: BLEU  \cite{papineni2002bleu}, ROUGE-L  \cite{lin-2004-rouge} and METEOR  \cite{denkowski-lavie-2014-meteor}, 
which were computed using the package released by \cite{sharma2017nlgeval}.

\subsection{Implementation Details}
We implemented our model in PyTorch 0.4.1  \cite{paszke2017pytorch} and train the model with a single Titan V.
The parameters setting and training techniques are described as follows.

We use the most frequent 20,000 words in training data to build the vocabulary for both the encoder and the decoder. The rest of the words are replaced as a generic $<$\textit{UNK}$>$ token. We set the word embedding size to 300, and initialize it by the pre-trained GloVe word vectors with 300 dimensions  \cite{pennington-etal-2014-glove}. The word embedding of words which are not included in Glove were initialized randomly. The NER, POS and answer position features are embedded to 16-dimensional vectors. The GRU hidden state sizes of both the encoder and the decoder are set to 512. We adopt dropout  \cite{srivastava2014dropout} with probability $p = 0.5$.

During training, We initialize model parameters randomly using a Gaussian distribution with Xavier scheme  \cite{glorot2010understanding}. The coverage mechanism with coverage loss weighted to $\lambda = 0.1$ (as described in \ref{cov}) is added. The sum of the cross-entropy loss and the coverage loss is optimized with the gradient descent algorithm by Adam  \cite{adam} optimizer. We set the initial learning rate $\alpha = 0.001$, two momentum parameters $\beta_1$ = 0.9 and $\beta_2$ = 0.999 respectively, and $\epsilon = 10^{-8} $. The learning rate starts to be reduced by half after training for 3500 steps. Gradient clipping  \cite{pascanu2013difficulty} with range [−5, 5] is applied.
The distribution over the decoder vocabulary is truncated to a distribution over the top k possible words, where k equals to 500.
The mini-batch size for the update is set to 64 and the model is trained up to 10 epochs. 
We select the model that achieves the best BLEU score on the dev set.
During decoding, beam search is conducted with the beam size of 15. 
The $\beta$ in the uncertainty-aware beam search is set to 0.115. 
Decoding stops when all the candidates in beam search generate the $<$\textit{EOS}$>$ token.

\subsection{Comparison}
To demonstrate the performance improvement of the proposed model, we compare it with the following baselines:
\begin{itemize}
	\item \textbf{NQG++}
	 \cite{A-Preliminary-Study} proposed an attention-based seq2seq model with a feature-rich encoder to encode word, answer position, POS and NER tagging information to generate answer focused question. 
	
	\item \textbf{s2s-a-at-mp-gsa}
	 \cite{zhao2018paragraph} extended previous seq2seq attention model with a gated self-attention encoder capable of utilizing relevant information from paragraph-level context and a maxout pointer mechanism to alleviate the repetition problem. 
	 
 	\item \textbf{MPQG}
	 \cite{song2018unified} proposed a multi-perspective matching encoder to capture the interactions between the paragraph and the target answer.
	 
	\item \textbf{ASs2s}
	 \cite{kim2019improving} proposed an answer-separated seq2seq which treats the paragraph and the target answer separately	for better utilization of the information from both sides.
	
	\item \textbf{CGC-QG}
	 \cite{liu2019learning} proposed the clue guided copy network which is a seq2seq model with copy mechanism and a variety of novel components and techniques (such as a clue word predictor) to boost the performance of question generation. 
	 
\end{itemize}

To quantify the contribution of the different components of our model, we evaluate the following versions:
\begin{itemize}
	\item \textbf{gsa-WCM}. 
	In this variant, instead of using our gated paragraph-to-answer attention, we adopt the gated self-attention proposed by \cite{zhao2018paragraph}.
	
	\item \textbf{EAN}. This model variant only considers the EAN capable of effectively utilizing of the information of the paragraph and the extended answer.
	
	\item \textbf{EAN-WCM}. To verify the effectiveness of WCM, this model variant removes UBS for the comparison with EAN.
	
	\item \textbf{EAN-WCM-UBS}. This is the complete version of our proposed model, which can be compared with EAN-WCM to verify the effectiveness of UBS.
\end{itemize}

\definecolor{antiquebrass}{rgb}{0.8, 0.58, 0.46}
\begin{table}[t]
	\centering
	\caption{Comparison of Generated Questions}
	\label{p2p_vs_p2s}
  \begin{threeparttable}
	\begin{tabular}{|p{0.9\columnwidth}|}
		\toprule
		\textbf{Paragraph:} the common pattern comes from john wesley , who wrote that `` there is no liturgy in the world , either in ancient or modern language , which breathes more of a solid , scriptural , rational piety , than the common prayer of the church of england . '' \textcolor{antiquebrass}{when the methodists in america were separated from the church of england , john wesley himself provided a revised version of \underline{the book of common prayer} called the sunday service of the methodists in north america .} wesley 's sunday service has shaped the official liturgies of the methodists ever since .\\
		\\
		\textbf{gsa-WCM:} what book did john wesley himself a revised version of ?\\
		\textbf{EAN-WCM:} the sunday service of the methodists in north america was a revised version of what ? \\
		\textbf{Ground truth:} the sunday service of the methodists in north america was a revised version of what book ? \\
		\bottomrule
	\end{tabular}
	\begin{tablenotes}
		\footnotesize
		\item The extended answer is colored, and the underlined words are the target answers.
	\end{tablenotes}
  \end{threeparttable}
\end{table}

\subsection{Results and Analysis}
The comparison results are given in Table~\ref{tab_scores}, our model EAN-WCM-UBS achieves the best results in all metrics.
We also demonstrate the effectiveness of our proposed components from different aspects.

\noindent \textbf{Extended Answer-aware Paragraph Encoder} 
To demonstrate the extended answer-aware encoder can effectively capture more relevant contextual information, 
we do a case study to compare the performance of our gated paragraph-to-answer attention and the gated self-attention proposed in \cite{zhao2018paragraph}. 
Table~\ref{p2p_vs_p2s} provides an example of questions generated by our model equipped with the gated self-attention and the gated paragraph-to-answer attention respectively. 
We can see our extended answer-aware network is capable of effectively utilizing more relevant information.

\begin{figure}[t]
	\centerline{\includegraphics[width = 0.95\columnwidth]{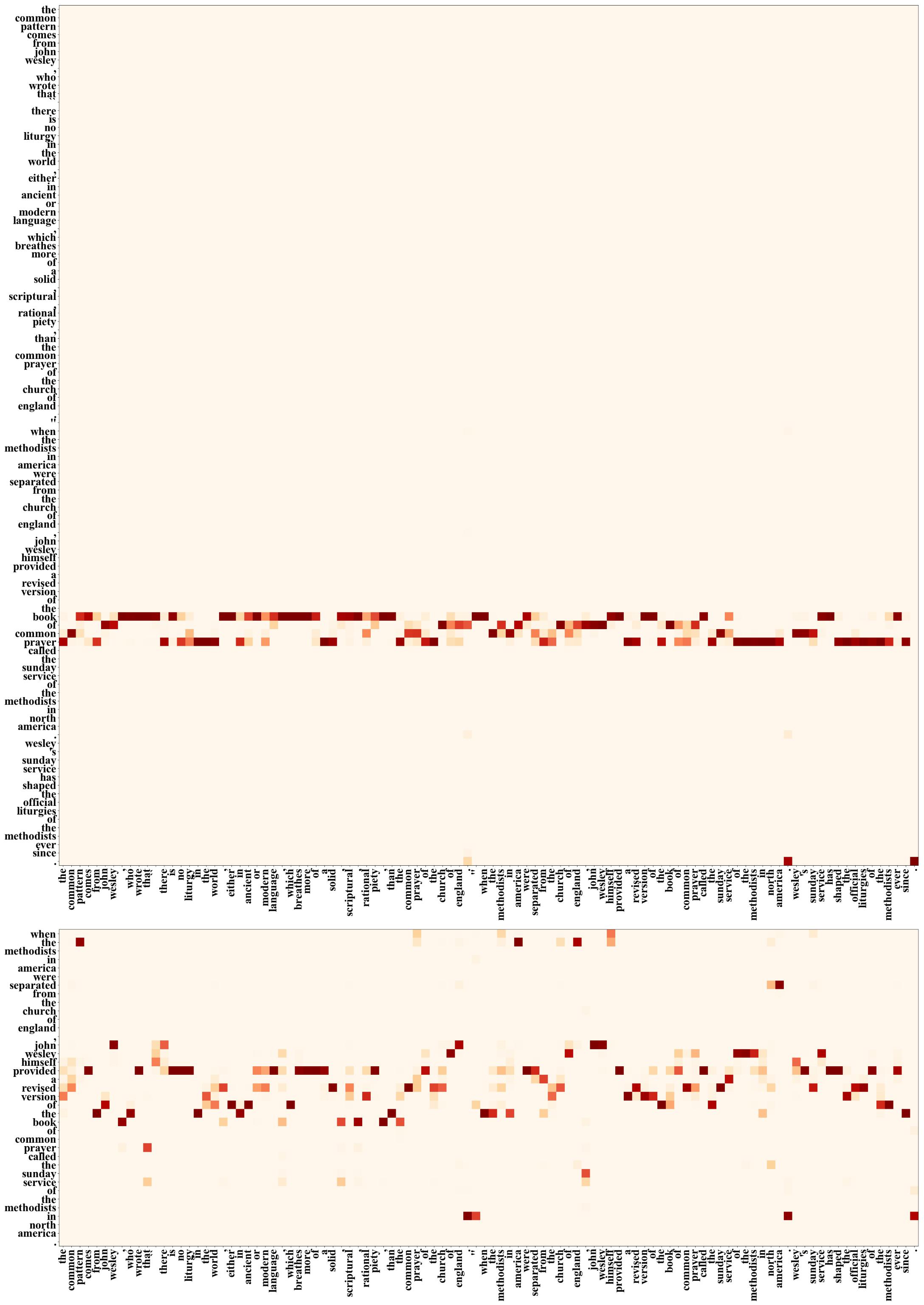}}
	\caption{Attention heatmaps:
		each column represents the attention weight vector.
		Above is the heatmap of the self-attention, and the heatmap of the paragraph-to-answer attention is drawn below.}
	\label{p2p_p2s_heatmaps}
\end{figure}

To further compare the paragraph-to-answer attention and the self-attention, we draw the heatmaps of both of the attention weight matrices in Figure~\ref{p2p_p2s_heatmaps} corresponding to the example in Table~\ref{p2p_vs_p2s}. 
We can see the attention distribution of the gated self-attention model always concentrates on the target answer words, while the paragraph-to-answer attention model results in capturing more relevant contextual information in the extended answer.

\begin{figure}[t]
	\centerline{\includegraphics[width = 0.95\columnwidth]{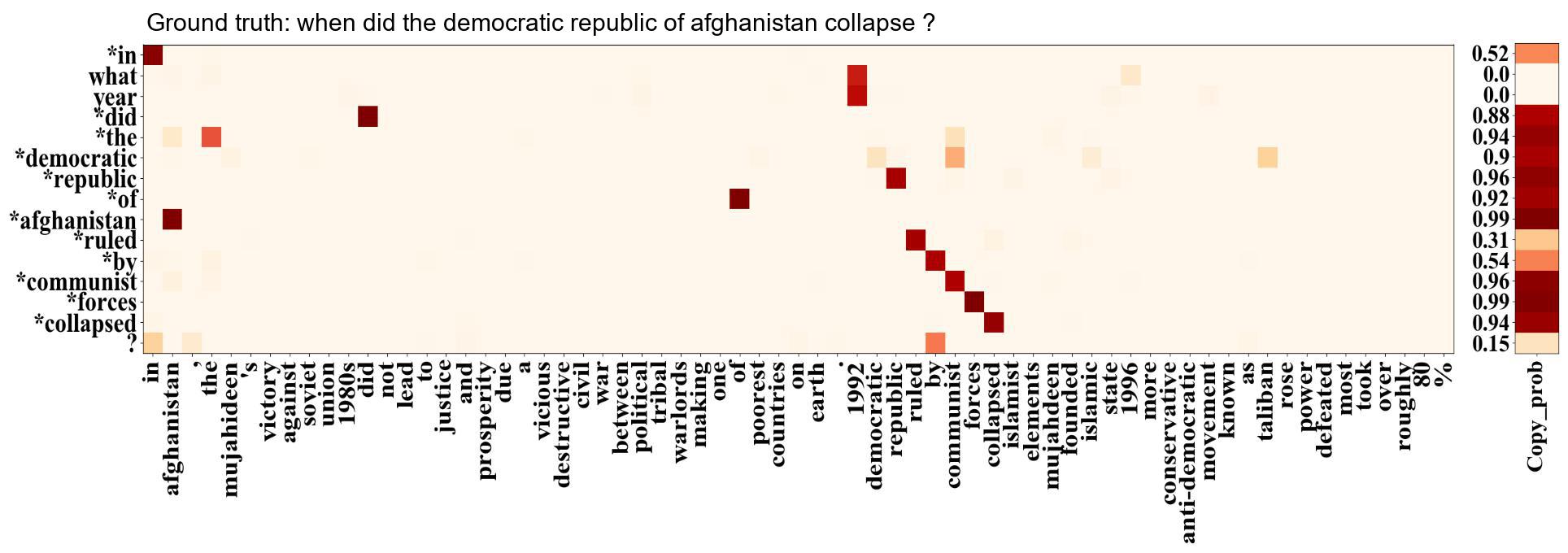}}
	\caption{Uncertainty-aware copy heatmap: each row represents a copy distribution and a corresponding copy probability decoded by model EAN+WCM+UBS.
		The words marked with * are words copied from input paragraph.
	}
	\label{ie}
\end{figure}

\noindent \textbf{Word-based Coverage Mechanism} To demonstrate the effectiveness of WCM can reduce repetitions, 
we show word duplication rates of the questions generated from various models in Table~\ref{dr}. 
Word duplication rate is computed by taking a ratio of the number of words appearing more than once over the total word counts. 
As shown in Table~\ref{dr}, the models trained with WCM (gsa-WCM, EAN-WCM, EAN-WCM-UBS) achieve much lower repetitions than the model trained without WCM (EAN, EAN-maxout, EAN-CM), 
where EAN-maxout and EAN-CM refer to EAN equipped with the maxout pointer mechanism \cite{zhao2018paragraph} and coverage mechanism \cite{see-etal-2017-get} respectively.
Besides, our UBS can further reduce undesirable repetitions, which can also be seen in the comparison between Figure~\ref{nie} and Figure~\ref{ie}.

\begin{table}[h]
	\centering
	\caption{Word duplication rates}
	\label{dr}
	\begin{tabular}{c|c}
		\toprule
		Models  			& Duplication Rate (\%) \\
		\midrule
		EAN	& 7.5		 \\
		EAN-maxout	& 7.2		 \\
		EAN-CM	& 6.38		 \\
		gsa-WCM	& 6.15		 \\
		EAN-WCM	& 6.1		 \\
		EAN-WCM-UBS	& \textbf{5.81}		 \\
		\midrule
		Ground Truth	& 3.59		 \\
		\bottomrule
	\end{tabular}
\end{table}

\noindent \textbf{Uncertainty-aware Beam Search}
We compute the BLEU score for our model where the $\beta$ has different values in the UBS and provide the results in Table~\ref{beta}. 
Sensitivity analysis on $\beta$ indicates that our method can improve the model performance in a certain range of $\beta$ values.

\begin{table}[h]
	\centering
	\caption{BLEU against $\beta$}
	\label{beta}
		\begin{tabular}{c|c|c|c|c|c|c}
			\toprule
			$\beta$ & 0 & 0.1 & 0.115 & 0.2 & 0.3 & 0.4 \\\midrule
			BLEU    & 17.41 & 17.53 & 17.56 & 17.51 & 17.38 & 17.19 \\
			\bottomrule
		\end{tabular}
\end{table}

To demonstrate the effectiveness of our proposed method can alleviate the inconsistent phenomenon between the copy probability and the copy distribution, 
we have compared the decoder copy heatmap Figure~\ref{ie} decoded with UBS with Figure~\ref{nie}. 
As shown in Figure~\ref{ie}, the decoder has more peaked attention distributions when it chooses to copy words from the input paragraph, which results in generating more accurate questions.

\section{Conclusion and Future Work}
In this paper, we propose an extended answer-aware network which is trained with word-based coverage mechanism and decodes with uncertainty-aware beam search.
Our EAN model can effectively capture more relevant contextual information in the extended answer using the gated paragraph-to-answer attention.
To reduce repetition, we design a WCM to penalize repeatedly attending to the same words at different time-steps.
Besides, we further observe that the inconsistent phenomenon between the copy probability and the copy distribution and design an UBS for decoding questions with less uncertainty. 
The experimental results on the SQuAD dataset show our method outperforms the baselines in QG task.

In future work, we would like to study the effectiveness of UBS for other language generation tasks which are benefited from the copy mechanism, such as neural machine translation and text summarization.

\bibliographystyle{aaai}
\bibliography{ref}

\end{document}